\documentclass[10pt,twocolumn,letterpaper]{article}

\usepackage{cvpr}
\usepackage{times}
\usepackage{epsfig}
\usepackage{graphicx}
\usepackage{amsmath}
\usepackage{amssymb}
\usepackage{multirow}
\usepackage{subfigure}

\usepackage[breaklinks=true,bookmarks=false]{hyperref}

\cvprfinalcopy 

\def\cvprPaperID{****} 

\begin{document}

\title{Bag of Tricks and A Strong Baseline for Deep Person Re-identification}
\newcommand*{\affaddr}[1]{#1}
\newcommand*{\affmark}[1][*]{\textsuperscript{#1}}
\author{Hao Luo\affmark[1]\footnotemark[1], Youzhi Gu\affmark[1]\footnotemark[1], Xingyu Liao\affmark[2]\footnotemark[1], Shenqi Lai\affmark[3], Wei Jiang\affmark[1]\\
\affaddr{\affmark[1]Zhejiang University},
\affaddr{\affmark[2]Chinese Academy of Sciences},
\affaddr{\affmark[3]Xi'an Jiaotong University} \\
{  \tt\small\{haoluocsc,gu$\_$youzhi,jiangwei$\_$zju\}@zju.edu.cn }
{ \tt\small  randall@mail.ustc.edu.cn }
{ \tt\small  laishenqi@stu.xjtu.edu.cn }
}

\maketitle
\thispagestyle{empty}
\newcommand\blfootnote[1]{%
\begingroup
\renewcommand\thefootnote{}\footnote{#1}%
\addtocounter{footnote}{-1}%
\endgroup
}

\renewcommand{\thefootnote}{\fnsymbol{footnote}}
\footnotetext[1]{Equal contributions. This work was partially done when Hao Luo and Xingyu Liao were interns at Megvii Inc.}


\begin{abstract}
This paper explores a simple and efficient baseline for person re-identification (ReID).
Person re-identification (ReID) with deep neural networks has made progress and achieved high performance in recent years.
However, many state-of-the-arts methods design complex network structure and concatenate multi-branch features.
In the literature, some effective training tricks are briefly appeared in several papers or source codes.
This paper will collect and evaluate these effective training tricks in person ReID.
By combining these tricks together, the model achieves 94.5\% rank-1 and 85.9\% mAP on Market1501 with only using global features.
Our codes and models are available at \href{https://github.com/michuanhaohao/reid-strong-baseline}{https://github.com/michuanhaohao/reid-strong-baseline}
\end{abstract}

\section{Introduction}

Person re-identification (ReID) with deep neural networks has made progress and achieved high performance in recent years.
However, many state-of-the-arts methods design complex network structure and concatenate multi-branch features.
In the literature, some effective training tricks or refinements are briefly appeared in several papers or source codes.
This paper will collect and evaluate such effective training tricks in person ReID.
With involved in all training tricks, ResNet50 reaches 94.5\% rank-1 accuracy and 85.9\% mAP on Market1501 \cite{zheng2015scalable}.
It is worth mentioning that it achieves such surprising performance with global features of the model.

\begin{figure}
\centering
\subfigure[Market1501]{
\begin{minipage}[b]{0.4\textwidth}
\includegraphics[width=1\textwidth]{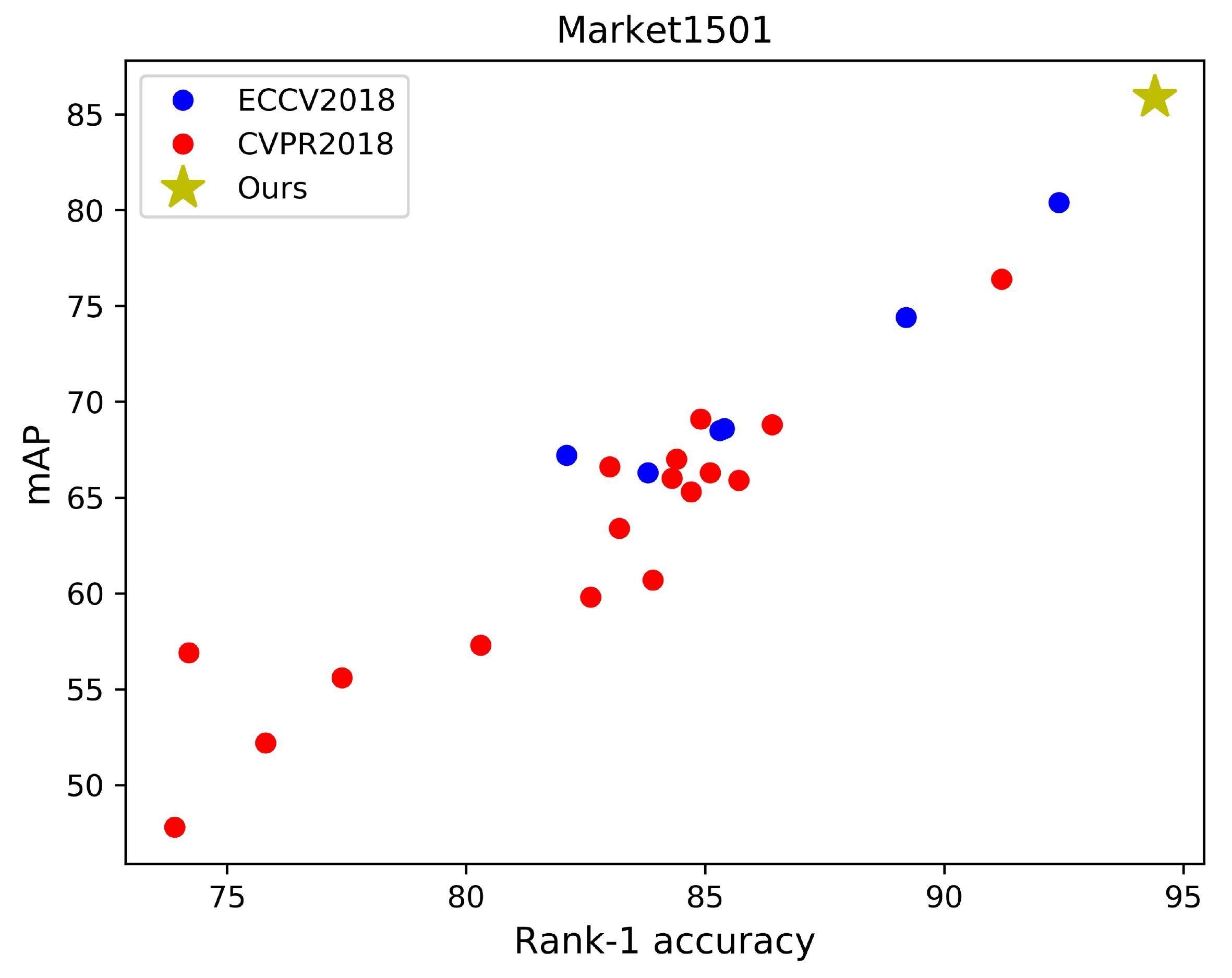}
\end{minipage}
}
\centering
\subfigure[DukeMTMC-reID]{
\begin{minipage}[b]{0.4\textwidth}
\includegraphics[width=1\textwidth]{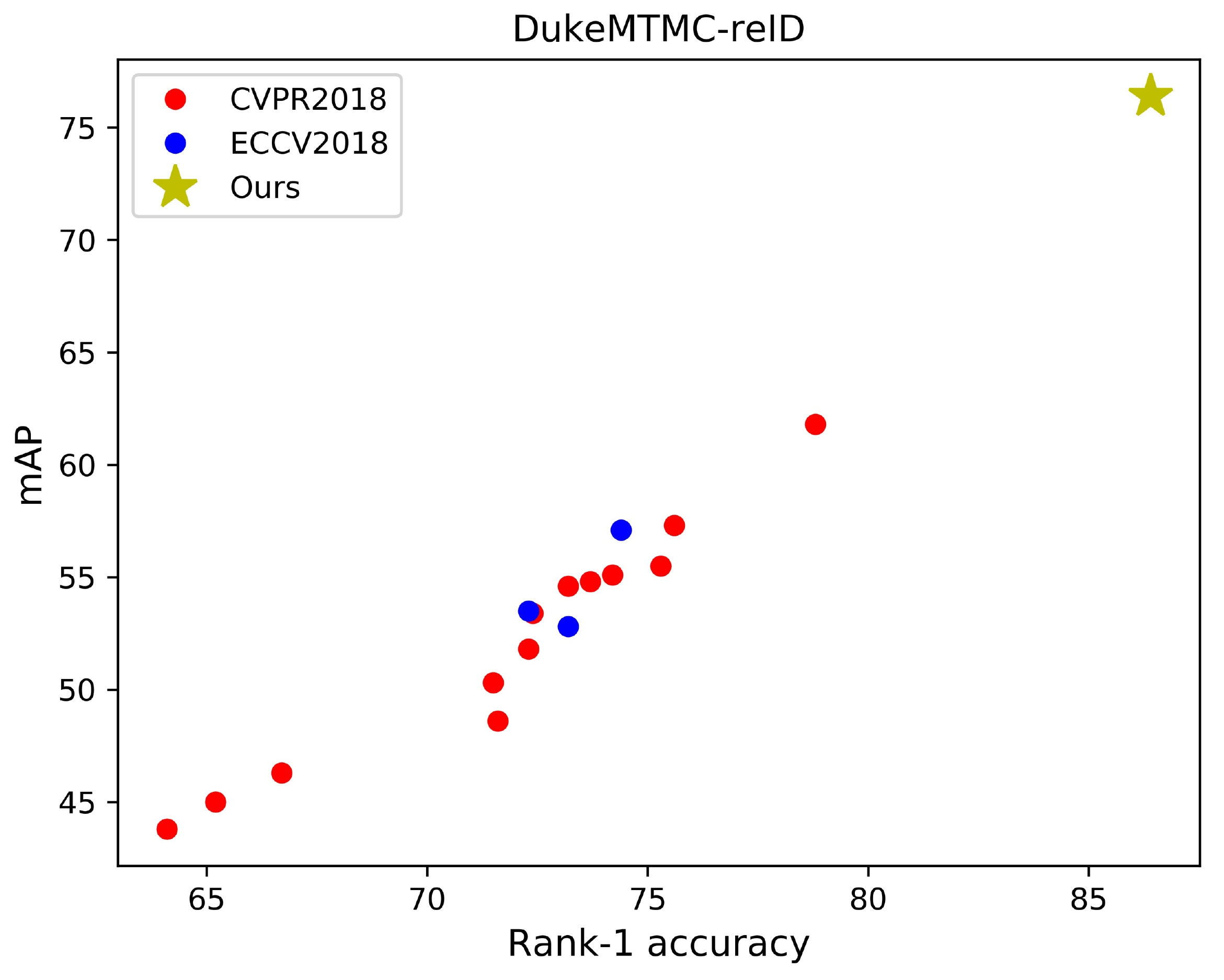}
\end{minipage}
}\caption{The performance of different baselines on Market1501 and DukeMTMC-reID datasets. We compare our strong baseline with other baselines published in CVPR2018 and ECCV2018.}\label{fig:baseline}
\end{figure}

For comparison, we surveyed articles published at ECCV2018 and CVPR2018 of the past year.
As shown in Fig. \ref{fig:baseline}, most of previous works were expanded on poor baselines.
On Market1501, only two baselines in 23 baselines surpassed 90\% rank-1 accuracy.
The rank-1 accuracies of four baselines even lower than 80\%.
On DukeMTMC-reID, all baselines did not surpass 80\% rank-1 accuracy or 65\% mAP.
We think a strong baseline is very important to promote the development of research.
Therefore, we modified the standard baseline with some training tricks to acquire a strong baseline.
The code of our strong baseline has been open sourced.

In addition, we also found that some works were unfairly compared with other state-of-the-arts methods.
Specifically, the improvements were mainly from training tricks rather than methods themselves.
But the training tricks were understated in the paper so that readers ignored them.
It would make the effectiveness of the method exaggerated.
We suggest that reviewers need to take into account these tricks when commenting academic papers.

Apart from aforementioned reasons, another consideration is that the industry prefers to simple and effective models rather than concatenating lots of local features in the inference stage.
In pursuit of high accuracy, researchers in the academic always combine several local features or utilize the semantic information from pose estimation or segmentation models.
Such methods bring too much extra consumption.
Large features also greatly reduce the speed of retrieval process.
Thus, we hope to use some tricks to improve the ability of the ReID model and only use global features to achieve high performance.
The purposes of this paper are summarized as follow:
\begin{itemize}
  \item We surveyed many works published on top conferences and found most of them were expanded on poor baselines.
  \item For the academia, we hope to provide a strong baseline for researchers to achieve higher accuracies in person ReID.
  \item For the community, we hope to give reviewers some references that what tricks will affect the performance of the ReID model. We suggest that when comparing the performance of the different methods, reviewers need to take these tricks into account.
  \item For the industry, we hope to provide some effective tricks to acquire better models without too much extra consumption.
\end{itemize}

Fortunately, a lot of effective training tricks have been present in some papers or open-sourced projects.
We collect many tricks and evaluate each of them on ReID datasets.
After a lot of experiments, we choose six tricks to introduce in this paper.
Some of them were designed or modified by us.
We add these tricks into a widely used baseline to get our modified baseline, which achieves 94.5\% rank-1 and 85.9\% mAP on Market1501.
Moreover, we found different works choose different image sizes and numbers of batch size, as a supplement, we also explore their impacts on model performance.
In summary, the contributions of this paper are concluded as follow:
\begin{itemize}
  \item We collect some effective training tricks for person ReID. Among them, we design a new neck structure named as BNNeck. In addition, we evaluate the improvements from each trick on two widely used datasets.
  \item We provide a strong ReID baseline, which achieves 94.5\% and 85.9\% mAP on Market1501. It is worth mentioned that the results are obtained with global features provided by ResNet50 backbone. To our best knowledge, it is the best performance acquired by global features in person ReID.
  \item As a supplement, we evaluate the influences of the image size and the number of batch size on the performance of ReID models.
\end{itemize}

\begin{figure*}
\centering
\subfigure[The pipeline of the standard baseline.]{
\begin{minipage}[b]{0.6\textwidth}
\includegraphics[width=1\textwidth]{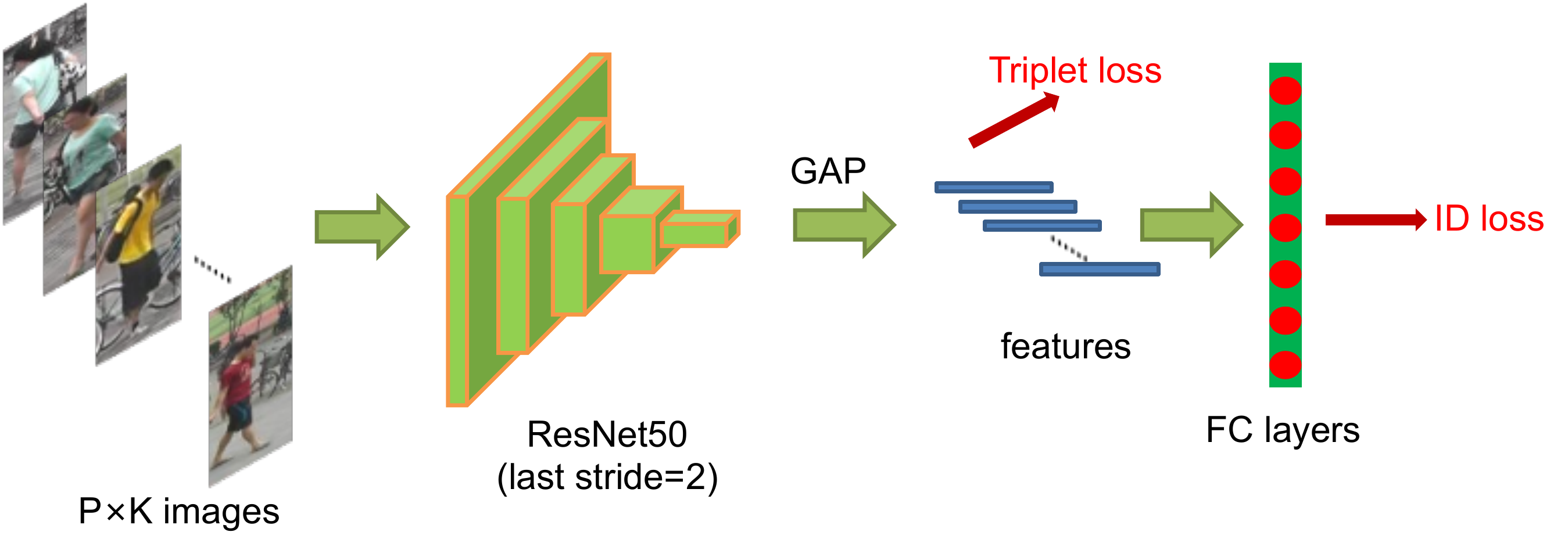}
\end{minipage}
}
\centering
\subfigure[The pipeline of our modified baseline.]{
\begin{minipage}[b]{0.8\textwidth}
\includegraphics[width=1\textwidth]{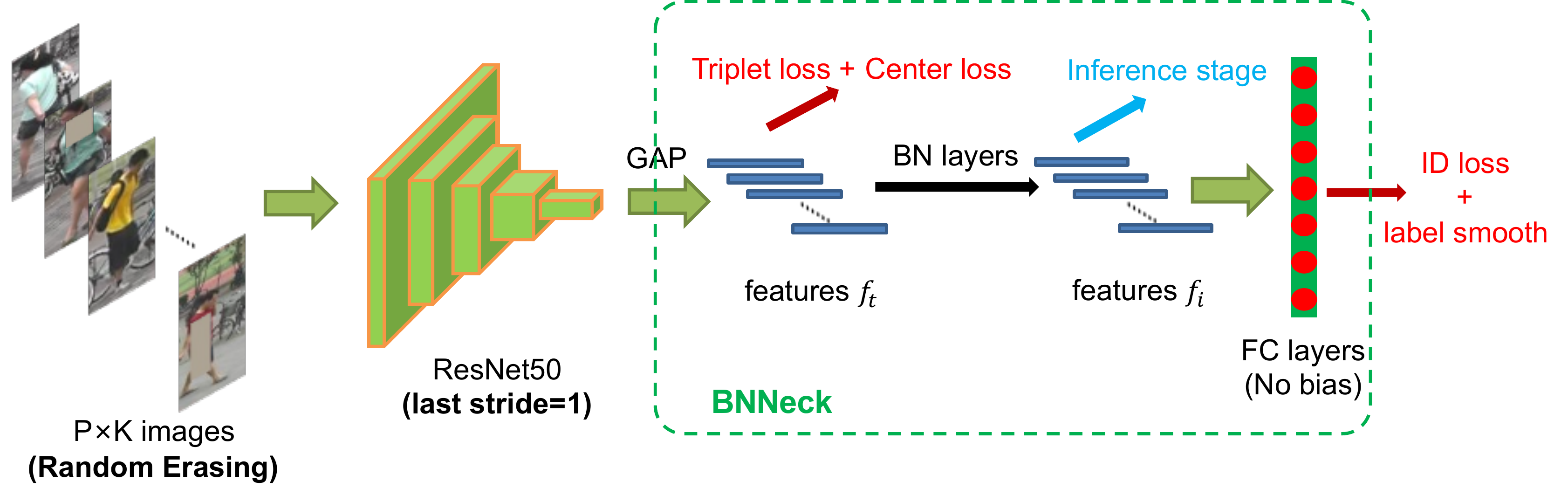}
\end{minipage}
}\caption{The pipelines of the standard baseline and our modified baseline.}\label{fig:arc}
\end{figure*}

\section{Standard Baseline}\label{SB}
We follow a widely used open-source\footnote[2]{\href{https://github.com/Cysu/open-reid}{https://github.com/Cysu/open-reid}} as our standard baseline.
The backbone of the standard baseline is ResNet50 \cite{he2016deep}.
During the training stage, the pipeline includes following steps:
\begin{enumerate}
  \item We initialize the ResNet50 with pre-trained parameters on ImageNet and change the dimension of the fully connected layer to $N$. $N$ denotes the number of identities in the training dataset.
  \item We randomly sample $P$ identities and $K$ images of per person to constitute a training batch. Finally the batch size equals to $B = P \times K$. In this paper, we set $P=16$ and $K=4$.
  \item We resize each image into $256\times 128$ pixels and pad the resized image 10 pixels with zero values. Then randomly crop it into a $256\times 128$ rectangular image.
  \item Each image is flipped horizontally with 0.5 probability.
  \item Each image is decoded into 32-bit floating point raw pixel values in $[0,1]$. Then we normalize RGB channels by subtracting 0.485, 0.456, 0.406 and dividing by 0.229, 0.224, 0.225, respectively.
  \item The model outputs ReID features $f$ and ID prediction logits $p$.
  \item ReID features $f$ is used to calculate triplet loss \cite{hermans2017defense}. ID prediction logits $p$ is used to calculated cross entropy loss. The margin $m$ of triplet loss is set to be 0.3.
  \item Adam method is adopted to optimize the model. The initial learning rate is set to be 0.00035 and is decreased by 0.1 at the 40th epoch and 70th epoch respectively. Totally there are 120 training epochs.
\end{enumerate}

\section{Training Tricks}
This section will introduce some effective training tricks in person ReID.
Most of such tricks can be expanded on the standard baseline without changing the model architecture.
The Fig. \ref{fig:arc} (b) shows training strategies and the model architecture appeared in this section.

\subsection{Warmup Learning Rate}

\begin{figure}[htb]
\centering
\includegraphics[width=.99\linewidth]{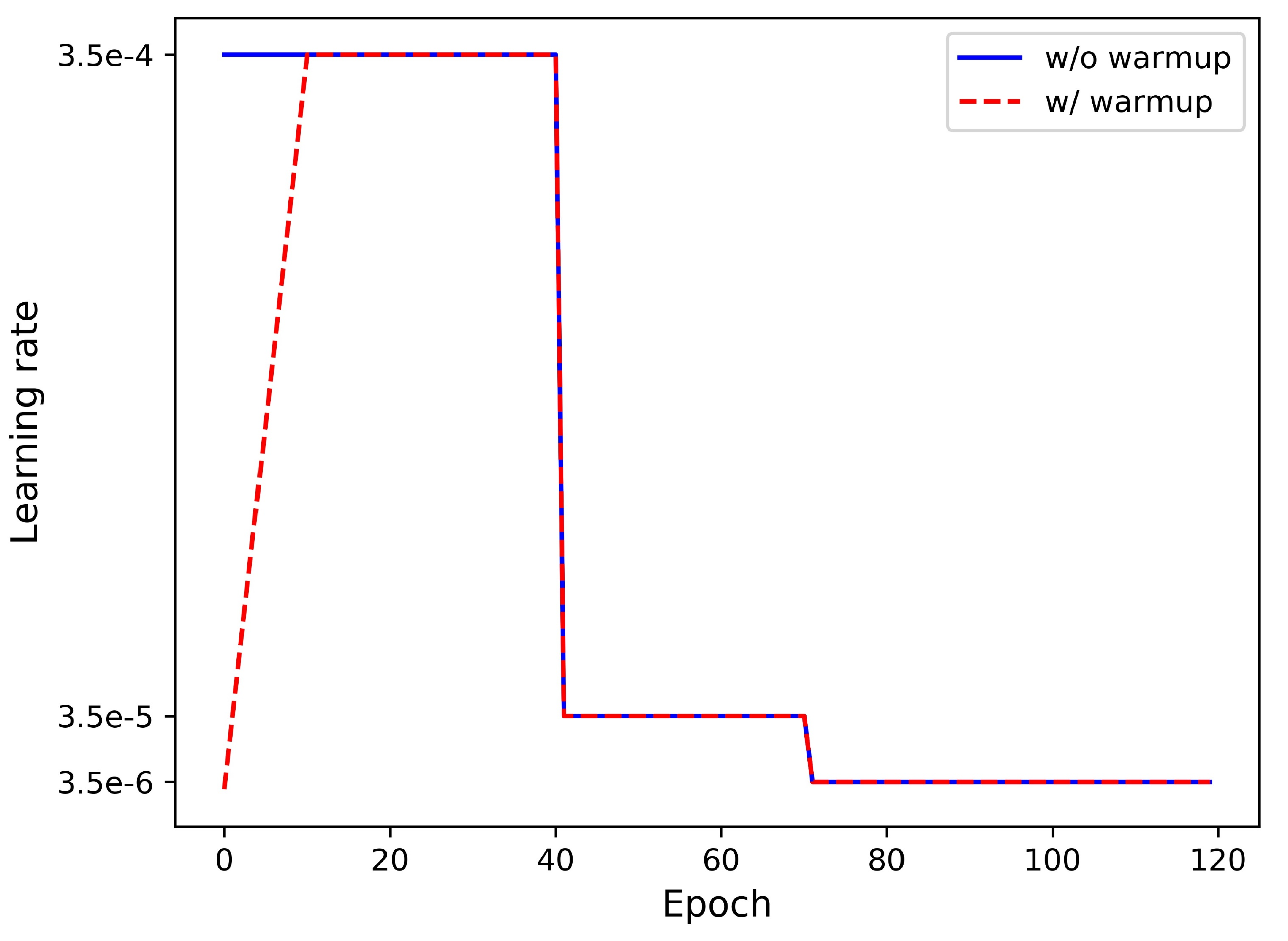}
\caption{Comparison of learning rate schedules. With warmup strategy, the learning rate is linearly increased in the first 10 epochs.}
\label{fig:lr}
\end{figure}

Learning rate has a great impact for the performance of a ReID model.
Standard baseline is initially trained with a large and constant learning rate.
In \cite{fan2019spherereid}, a warmup strategy is applied to bootstrap the network for better performance.
In practice, As shown in Fig. \ref{fig:lr}, we spent 10 epochs linearly increasing the learning rate from $3.5\times 10^{-5}$ to $3.5\times 10^{-4}$.
Then, the learning rate is decayed to $3.5\times 10^{-5}$ and $3.5\times 10^{-6}$ at 40th epoch and 70th epoch respectively.
The learning rate $lr(t)$ at epoch $t$ is compute as;
\begin{equation}
lr(t)=\left\{\begin{array}{ll}
{3.5 \times 10^{-5} \times \frac{t}{10} }  & {\text { if } t \leq 10} \\
{3.5 \times 10^{-4}} & {\text { if } 10< t \leq 40} \\
{3.5 \times 10^{-5}}  & {\text { if } 40< t \leq 70} \\
{3.5 \times 10^{-6}} & {\text { if } 70< t \leq 120}
\end{array}\right.
\end{equation}

\subsection{Random Erasing Augmentation}
In person ReID, persons in the images are sometimes occluded by other objects.
To address the occlusion problem and improve the generalization ability of ReID models, Zhong \emph{et al.} \cite{zhong2017random} proposed a new data augmentation approach named as Random Erasing Augmentation (REA).
In practice, for an image $I$ in a mini-batch, the probability of it undergoing Random Erasing is $p_e$, and the probability of it being kept unchanged is $1-p_e$.
Then, REA randomly selects a rectangle region $I_e$ with size $(W_e, H_e)$ in image $I$, and erases its pixels with random values.
Assuming the area of image $I$ and region $I_e$ are $S = W \times H$ and $S_e = W_e \times H_e$ respectively, we denote $r_e = \frac{S_e}{S}$ as the area ratio of erasing rectangle region.
In addition, the aspect ratio of region $I_e$ is randomly initialized between $r_1$ and $r_2$.
To determine a unique region, REA randomly initializes a point $\mathcal{P} = \left(x_{e} , y_{e} \right)$.
If $x_{e} + W_{e} \leq W$ and $y_{e} + H_{e} \leq H$, we set the region, $I_{e} = \left( x_{e}, y_{e},x_{e}+W_{e},y_{e}+H_{e} \right)$, as the selected rectangle region.
Otherwise we repeat the above process until an appropriate $I_e$ is selected.
With the selected erasing region $I_e$, each pixel in $I_e$ is assigned to the mean value of image $I$, respectively.

\begin{figure}[tb]
\centering
\includegraphics[width=.99\linewidth]{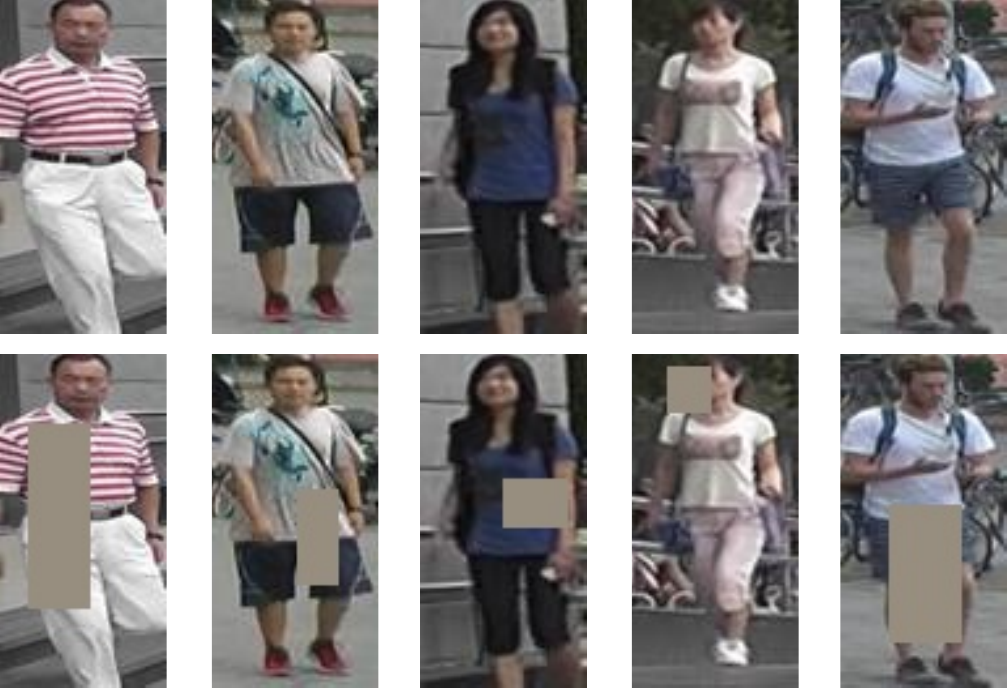}
\caption{Sampled examples of random erasing augmentation. The first row shows five original training images. The processed images are presented in the second low.}
\label{fig:rea}
\end{figure}

In this study, we set hyper-parameters to $p=0.5, 0.02<S_e<0.4, r_1=0.3, r_2=3.33$, respectively.
Some examples are shown in Fig. \ref{fig:rea}.

\subsection{Label Smoothing}
ID Embedding (IDE) \cite{zheng2018discriminatively} network is a basic baseline in person ReID.
The last layer of IDE, which outputs the ID prediction logits of images, is a fully-connected layer with a hidden size being equal to numbers of persons $N$.
Given an image, we denote $y$ as truth ID label and $p_i$ as ID prediction logits of class $i$.
The cross entropy loss is computed as:
\begin{equation}
L(ID)= \sum_{i=1}^{N}-q_{i} \log \left(p_i\right) \left\{\begin{array}{ll}
q_i = 0, y \ne i \\
q_i = 1, y=i
\end{array}\right.
\end{equation}
Because the category of the classification is determined by the person ID, we call such loss function as ID loss in this paper.

Nevertheless, person ReID can be regard as one-shot learning task because person IDs of the testing set have not appeared in the training set.
So it is pretty important to prevent the ReID model from overfitting training IDs.
Label smoothing (LS) proposed in \cite{szegedy2016rethinking} is a widely used method to prevent overfitting for a classification task.
It changes the construction of $q_i$ to:
\begin{equation}\label{LS}
  q_{i}=\left\{\begin{array}{ll}
  {1- \frac{N-1}{N} \varepsilon} & {\text { if } i=y} \\
  {\varepsilon /N} & {\text { otherwise, }}\end{array}\right.
\end{equation}
where $\varepsilon$ is a small constant to encourage the model to be less confident on the training set.
In this study, $\varepsilon$ is set to be $0.1$.
When the training set is not very large, LS can significantly improve the performance of the model.

\subsection{Last Stride}
Higher spatial resolution always enriches the granularity of feature.
In \cite{sun2018beyond}, Sun \emph{et al.} removed the last spatial down-sampling operation in the backbone network to increase the size of the feature map.
For convenience, we denote the last spatial down-sampling operation in the backbone network as last stride.
The last stride of ResNet50 is set to be 2.
When fed into a image of $256 \times 128$ size, the backbone of ResNet50 outputs a feature map with the spatial size of $8 \times 4$.
If change last stride from 2 to 1, we can get a feature map with higher spatial size ($16 \times 8$).
This manipulation only increases very light computation cost and does not involve extra training parameters.
However, higher spatial resolution brings significant improvement.

\subsection{BNNeck}

\begin{figure}[b]
\centering
\subfigure[The neck of the standard baseline.]{
\begin{minipage}[b]{0.25\textwidth}
\includegraphics[width=1\textwidth]{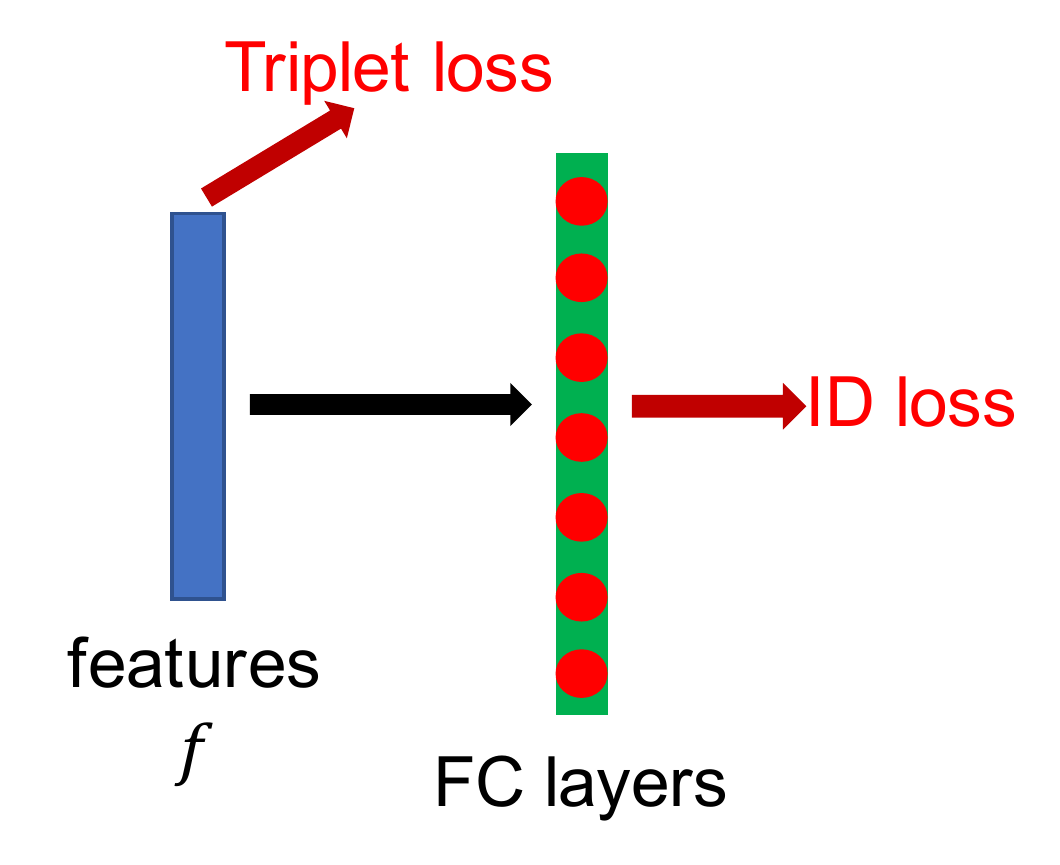}
\end{minipage}
}
\centering
\subfigure[Our designed BNNeck. In the inference stage, we choose $f_i$ following the BN layer to do the retrieval.]{
\begin{minipage}[b]{0.4\textwidth}
\includegraphics[width=1\textwidth]{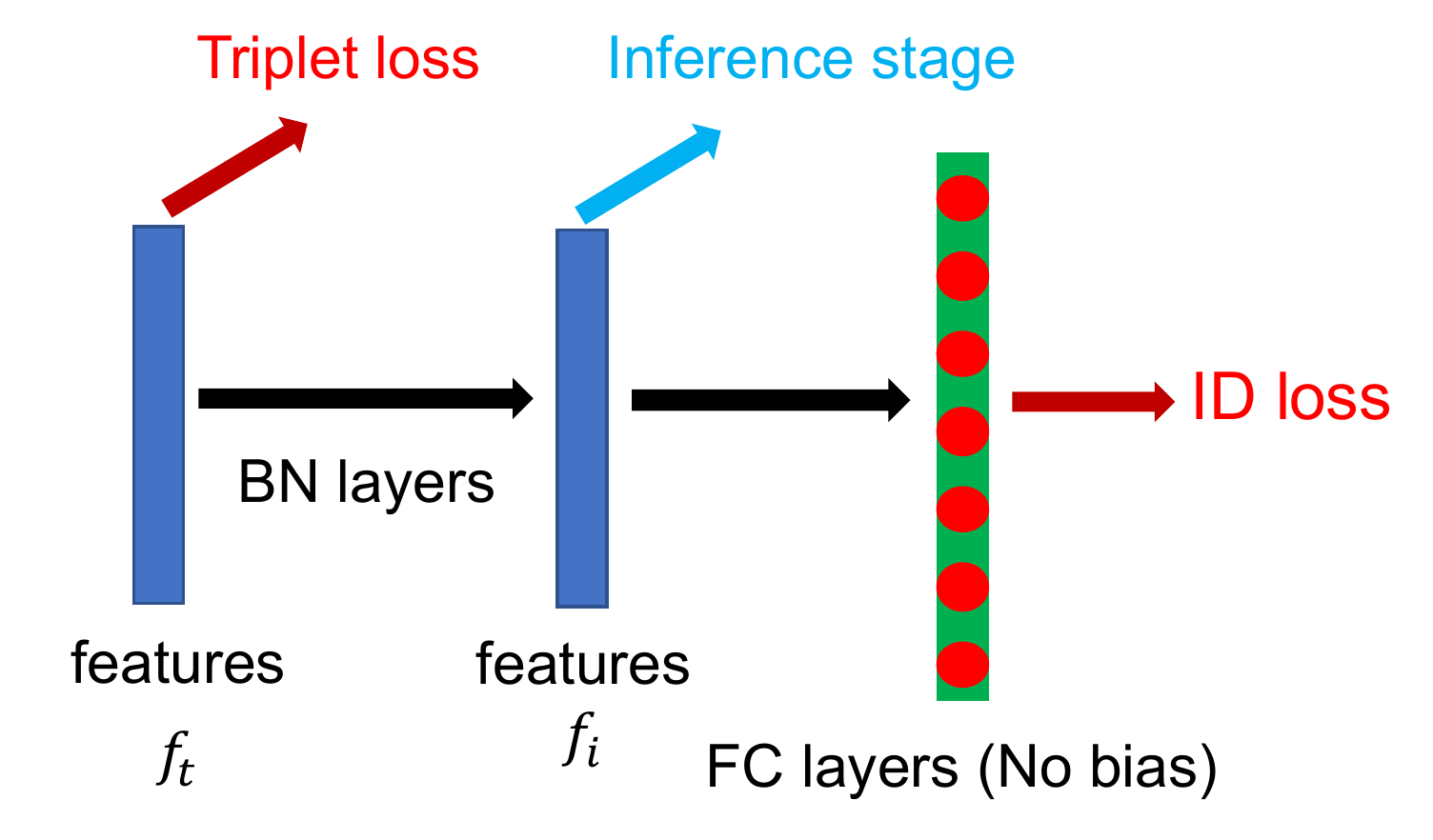}
\end{minipage}
}\caption{Comparison between standard neck and our designed BNNeck.}\label{fig:bnneck}
\end{figure}

\begin{figure*}[htb]
\centering
\includegraphics[width=.99\linewidth]{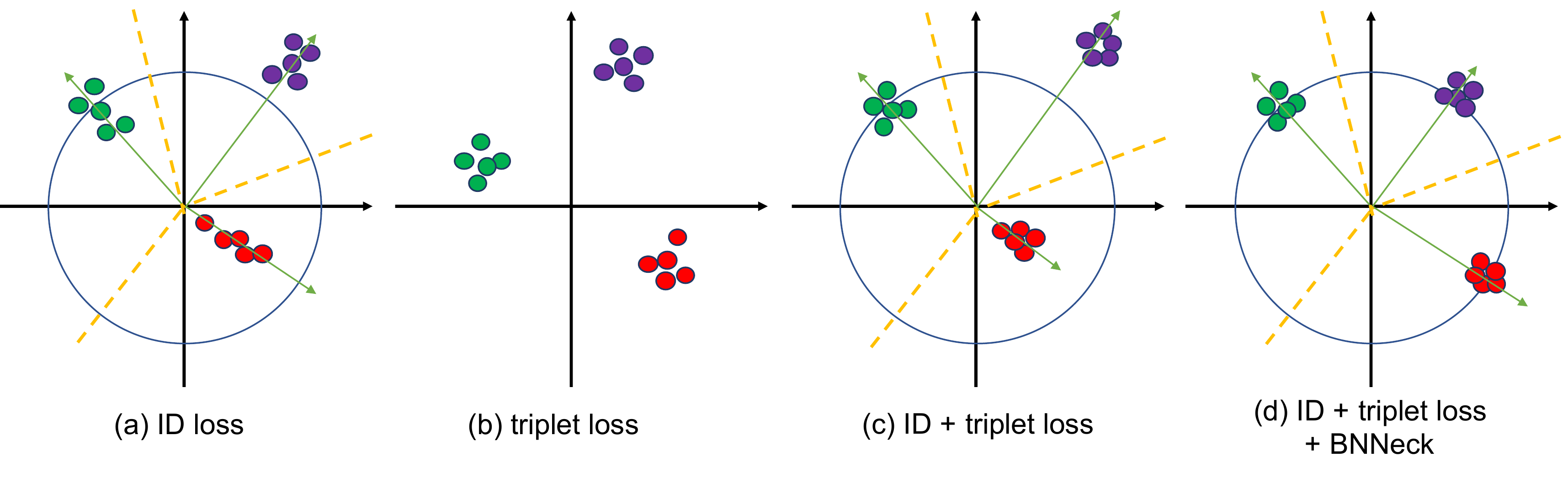}
\caption{Two-dimensional visualization of sample distribution in the embedding space supervised by (a) ID Loss, (b) Triplet Loss, (c) ID + triplet loss and (d) ID + triplet loss + BNNeck.
Points of different colors represent embedding features from different classes. The yellow dotted lines stand for the supposed classification hyperplanes.}
\label{fig:eb}
\end{figure*}

Most of works combined ID loss and triplet loss together to train ReID models.
As shown in Fig. \ref{fig:bnneck}(a), in the standard baseline, ID loss and triplet loss constrain the same feature $f$.
However, the targets of these two losses are inconsistent in the embedding space.

As shown in Fig. \ref{fig:eb}(a), ID loss constructs several hyperplanes to separate the embedding space into different sub-spaces.
The features of each class are distributed in different subspaces.
In this case, cosine distance is more suitable than Euclidean distance for the model optimized by ID loss in the inference stage.
On the other hand, as shown in \ref{fig:eb}(b), triplet loss enhances the intra-class compactness and inter-class separability in the Euclidean space.
Because triplet loss can not provide globally optimal constraint, inter-class distance sometimes is smaller than intra-class distance.
A widely used method is to combine ID loss and triplet loss to train the model together.
This approach let the model learn more discriminative features.
Nevertheless, for image pairs in the embedding space, ID loss mainly optimizes the cosine distances while triplet loss focuses on the Euclidean distances.
If we use these two losses to simultaneously optimize a feature vector, their goals may be inconsistent.
In the training process, a possible phenomenon is that one loss is reduced, while the other loss is oscillating or even increased.

To overcome the aforementioned problem, we design a structure named as BNNeck shown in Fig. \ref{fig:bnneck}(b).
BNNeck only adds a batch normalization (BN) layer after features (and before classifier FC layers).
The feature before the BN layer is denoted as $f_t$.
We let $f_t$ pass through a BN layer to acquire the normalized feature $f_i$.
In the training stage, $f_t$ and $f_i$ are used to compute triplet loss and ID loss, respectively.
Normalization balances each dimension of $f_i$.
The features are gaussianly distributed near the surface of the hypersphere.
This distribution makes the ID loss easier to converge.
In addition, BNNeck reduces the constraint of the ID loss on $f_t$.
Less constraint from ID loss leads to triplet loss easier to converge at the same time.
Thirdly, normalization keeps the compact distribution of features that belong to one same person.

Because the hypersphere is almost symmetric about the origin of the coordinate axis, another trick of BNNeck is removing the bias of classifier FC layer.
It constrains the classification hyperplanes to pass through the origin of the coordinate axis.
We initialize the FC layer with Kaiming initialization proposed in \cite{he2015delving}.

In the inference stage, we choose $f_i$ to do the person ReID task.
Cosine distance metric can achieve better performance than Euclidean distance metric.
Experimental results in Table. \ref{ablation} show that BNNeck can improve performance of the ReID model by a large margin.

\subsection{Center Loss}

Triplet loss is computed as:
\begin{equation}\label{eq:tri}
  L_{Tri} = [d_p - d_n + \alpha]_{+},
\end{equation}
where $d_p$ and $d_n$ are feature distances of positive pair and negative pair.
$\alpha$ is the margin of triplet loss, and $[z]_{+}$ equals to $max(z,0)$.
In this paper, $\alpha$ is set to $0.3$.
However, triplet loss only considers the difference between $d_p$ and $d_n$ and ignores the absolute values of them.
For instance, when $d_p=0.3, d_n=0.5$, the triplet loss is $0.1$.
For another case, when $d_p=1.3, d_n=1.5$, the triplet loss also is $0.1$.
Triplet loss is determined by two person IDs sampled randomly.
It is difficult to ensure that $d_p < d_n$ in the whole training dataset.

Center loss \cite{wen2016discriminative}, which simultaneously learns a center for deep features of each class and penalizes the distances between the deep features and their corresponding class centers, makes up for the drawbacks of the triplet loss.
The center loss function is formulated as:
\begin{equation}\label{center}
  \mathcal{L}_{C}=\frac{1}{2} \sum_{j=1}^{B}\left\|\boldsymbol{f}_{t_j}-\boldsymbol{c}_{y_{j}}\right\|_{2}^{2},
\end{equation}
where $y_{j}$ is the label of the $j$th image in a mini-batch.
$\boldsymbol{c}_{y_{j}}$ denotes the $y_i$th class center of deep features.
$B$ is the number of batch size.
The formulation effectively characterizes the intra-class variations.
Minimizing center loss increases intra-class compactness.
Our model totally includes three losses as follow:
\begin{equation}\label{loss}
  L = L_{ID} + L_{Triplet} + \beta L_{C}
\end{equation}
$\beta$ is the balanced weight of center loss.
In our experiments, $\beta$ is set to be $0.0005$.

\section{Experimental Results}
In this section, we will evaluate our models on Market1501 and DukeMTMC-reID \cite{ristani2016MTMC} datasets.
The Rank-1 accuracy and mean Average Precision (mAP) are reported as evaluation metrics.
We add tricks on the standard baseline successively and do not change any training settings.
The results of ablation studies present the performance boost from each trick.
In order to prevent being misled by overfitting, we also show the results of cross-domain experiments.

\subsection{Influences of Each Trick (Same domain)}

\renewcommand{\multirowsetup}{\centering}
\begin{table}[htb]\small
  \begin{center}
  \begin{tabular}{ l|cc|cc}
\hline
    			& \multicolumn{2}{c|}{Market1501} & \multicolumn{2}{c}{DukeMTMC}	 \\
  Model			& r = 1 	& mAP	&r = 1 	& mAP 	 \\
 	\hline
	\hline
Baseline-S		&87.7	&74.0	&79.7	&63.7			\\
+warmup         &88.7	&75.2	&80.6	&65.1           \\
+REA            &91.3	&79.3	&81.5	&68.3			\\
+LS	            &91.4	&80.3	&82.4	&69.3			\\
+stride=1	    &92.0	&81.7	&82.6	&70.6			\\
+BNNeck	        &94.1	&85.7	&86.2	&75.9		\\
+center loss    &94.5	&85.9	&86.4	&76.4		\\
\hline

  \end{tabular}
  \end{center}
  \caption{\label{ablation}The performance of different models is evaluated on Market1501 and DukeMTMC-reID datasets. Baseline-S stands for the standard baseline introduced in section \ref{SB}.}
\end{table}

The standard baseline introduced in section \ref{SB} achieves 87.7\% and 79.7\% rank-1 accuracies on Market1501 and DukeMTMC-reID, respectively.
The performance of standard baseline is similar with most of baselines reported in other papers.
Then, we add warmup strategy, random erasing augmentation, label smoothing, stride change, BNNeck and center loss to the model training process, one by one.
Our designed BNNeck boosts more performance than other tricks, especially on DukeMTMC-reID.
Finally, these tricks make baseline acquire 94.5\% rank-1 accuracy and 85.9\% mAP on Market1501.
On DukeMTMC-reID, it reaches 86.4\% rank-1 accuracy and 76.4\% mAP.
In other works, these training tricks boost the performance of the standard baseline by more than 10\% mAP.
In addition, to get such improvement, we only involve an extra BN layer and do not increase training time.

\subsection{Analysis of BNNeck}

\renewcommand{\multirowsetup}{\centering}
\begin{table}[htb]\small
  \begin{center}
  \begin{tabular}{ c|c|cc|cc}
\hline
    		&		       & \multicolumn{2}{c|}{Market1501} & \multicolumn{2}{c}{DukeMTMC}	 \\
  Feature   & Metric	   & r = 1 	& mAP	&r = 1 	& mAP 	 \\
 	\hline
	\hline
    $f$ (w/o BNNeck) & Euclidean    &92.0	&81.7	&82.6	&70.6 \\
    $f_t$   & Euclidean             &94.2	&85.5	&85.7	&74.4		\\
    $f_t$   & Cosine                &94.2	&85.7	&85.5	&74.6		\\
    $f_i$   & Euclidean	            &93.8	&83.7	&86.6	&73.0       \\
    $f_i$   & Cosine                &\textbf{94.1}	&\textbf{85.7}	&\textbf{86.2}	&\textbf{75.9}		\\

\hline

  \end{tabular}
  \end{center}
  \caption{\label{tab:bnneck}The ablation study of BNNeck. $f$ (w/o BNNeck) is baseline without BNNeck. BNNeck includes two features $f_t$ and $f_i$. We evaluate the performance of them with Euclidean distance and cosine distance, respectively.}
\end{table}

In this section, we evaluate the performance of two different features ($f_t$ and $f_i$) with Euclidean distance metric and cosine distance metric.
All models are trained without center loss in Table. \ref{tab:bnneck}.
We observe that cosine distance metric performs better than Euclidean distance metric for $f_t$.
Because ID loss directly constrains the features followed the BN layer, $f_i$ can be separated by several hyperplanes clearly.
The cosine distance can measure the angle between two feature vectors, so cosine distance metric is more suitable than Euclidean distance metric for $f_i$.
However, $f_t$ is close to triplet loss and is constrained by ID loss at the same time.
Two kinds of metrics achieve similar performance for $f_t$.

In overall, BNNeck significantly improve the performance of ReID models.
We choose $f_i$ with cosine distance metric to do the retrieval in the inference stage.

\subsection{Influences of Each Trick (Cross domain)}

\renewcommand{\multirowsetup}{\centering}
\begin{table}[b]\small
  \begin{center}
  \begin{tabular}{ l|cc|cc}
\hline
    			& \multicolumn{2}{c|}{M$\to$D} & \multicolumn{2}{c}{D$\to$M}	 \\
  Model			& r = 1 	& mAP	&r = 1 	& mAP 	 \\
 	\hline
	\hline
Baseline		&24.4	&12.9	&34.2	&14.5			\\
+warmup         &26.3	&14.1	&39.7	&17.4           \\
+REA            &21.5	&10.2	&32.5	&13.5			\\
+LS	            &23.2	&11.3	&36.5	&14.9			\\
+stride=1	    &23.1	&11.8	&37.1	&15.4			\\
+BNNeck	        &26.7	&15.2	&47.7	&21.6			\\
+center loss	&27.5	&15.0	&47.4	&21.4			\\
-REA            &41.4	&25.7	&54.3	&25.5			\\
\hline

  \end{tabular}
  \end{center}
  \caption{\label{cd}The performance of different models is evaluated on cross-domain datasets. M$\to$D means that we train the model on Market1501 and evaluate it on DukeMTMC-reID.}
\end{table}

To further explore effectiveness, we also present the results of cross-domain experiments in Table. \ref{cd}.
In overview, three tricks including warmup strategy, label smoothing and BNNeck significantly boost the cross-domain performance of ReID models.
Stride change and center loss seem to have no big impact on the performance.
However, REA does harm to models in cross-domain ReID task.
In particularly, when our modified baseline is trained without REA, it achieves 41.4\% and 54.3\% rank-1 accuracies on Market1501 and DukeMTMC-reID datasets, respectively.
Its performance surpass the ones of the standard baseline by a large margin.
We infer that REA masking the regions of training images lets the model learn more knowledge in the training domain.
It causes the model to perform worse in the testing domain.

\subsection{Comparison of State-of-the-Arts}

\renewcommand{\multirowsetup}{\centering}
\begin{table}[tb]\footnotesize
  \begin{center}
  \begin{tabular}{ ccc|cc|cc}
\hline
    		&		&       & \multicolumn{2}{c|}{Market1501} & \multicolumn{2}{c}{DukeMTMC}	 \\
  Type   & Method & $N_f$	   & r = 1 	& mAP	&r = 1 	& mAP 	 \\
 	\hline
	\hline
    \multirow{3}{1cm}{Pose-guided}& GLAD\cite{wei2017glad}        & 4  &89.9	&73.9	&-	&-		\\
                                & PIE \cite{zheng2017pose}          & 3  &87.7	&69.0	&79.8	&62.0		\\
                                & PSE \cite{Sarfraz_2018_CVPR}      & 3  &78.7	&56.0	&-	&-		\\
    \hline
    \multirow{2}{1cm}{Mask-guided}& SPReID \cite{kalayeh2018human}    & 5  & 92.5 & 81.3	& 84.4	&71.0		\\
                                & MaskReID \cite{qi2018maskreid}    & 3  &90.0	&75.3	&78.8	&61.9		\\
 \hline
    \multirow{6}{1cm}{Stripe-based}& AlignedReID \cite{zhang2017alignedreid}& 1 &90.6 &77.7 &81.2	&67.4 		\\
                                & SCPNet \cite{fan2018scpnet}       & 1  & 91.2	&75.2	&80.3	&62.6		\\
                                & PCB \cite{sun2018beyond}          & 6  & 93.8	&81.6	&83.3	&69.2		\\
                                & Pyramid\cite{zheng2018coarse}     & 1 & 92.8 &82.1	&-	&-		\\
                                & Pyramid\cite{zheng2018coarse}     & 21 & 95.7 &88.2	&89.0	&79.0		\\
                                & BFE\cite{dai2018batch}            & 2 & 94.5 &85.0	&88.7	&75.8		\\
   \hline
    \multirow{3}{1cm}{Attention-based}& Mancs \cite{wang2018mancs}        & 1 &93.1   &82.3	&84.9	&71.8		\\
                                & DuATM \cite{si2018dual}           & 1 & 91.4	& 76.6	&81.2	&62.3		\\
                                & HA-CNN \cite{li2018harmonious}    & 4 & 91.2	& 75.7	&80.5	&63.8		\\
    \hline
    \multirow{2}{1cm}{GAN-based}& Camstyle \cite{zhong2019camstyle} &  1 &88.1 	&68.7	&75.3	&53.5		\\
                                & PN-GAN \cite{qian2018pose}        &  9 &89.4 	&72.6	&73.6	&53.2		\\
    \hline
    \multirow{6}{1cm}{Global feature}& IDE \cite{zheng2018discriminatively}  & 1  & 79.5	& 59.9	& -	&-		\\
                                & SVDNet \cite{sun2017svdnet}  & 1  & 82.3	& 62.1	& 76.7	&56.8		\\
                                & TriNet\cite{hermans2017defense}  & 1  & 84.9	& 69.1	& -	& -		\\
                                & AWTL\cite{ristani2018features}  & 1  & 89.5	& 75.7	& 79.8	& 63.4		\\
                                & \textbf{Ours}    &  1 &\textbf{94.5}	&\textbf{85.9}	&\textbf{86.4}	&\textbf{76.4}		\\
                                & \textbf{Ours(RK)}&  1 &\textbf{95.4}	&\textbf{94.2}	&\textbf{90.3}	&\textbf{89.1}		\\
\hline

  \end{tabular}
  \end{center}
  \caption{\label{tab:sota}Comparison of state-or-the-arts methods. $N_f$ is the number of features used in the inference stage. RK stands for $k$-reciprocal re-ranking method \cite{zhong2017re}}
\end{table}

We compare out strong baseline with state-of-the-arts methods in Table. \ref{tab:sota}.
All methods have been divided into different types.
Pyramid\cite{zheng2018coarse} achieves surprising performance on two datasets.
However, it concatenates 21 local features of different scale.
If only utilizing the global feature, it obtains 92.8\% rank-1 accuracy and 82.1\% mAP on Market1501.
Ours strong baseline can reach 94.5\% rank-1 accuracy and 85.9\% mAP on Market1501.
BFE\cite{dai2018batch} obtains similar performance with our strong baseline.
But it combines features of two branches.
Throughout all methods that only use global features, our strong baseline beats AWTL\cite{ristani2018features} by more than 10\% mAP on both Market1501 and DukeMTMC-reID.
With $k$-reciprocal re-ranking method to boost the performance, our method reaches 94.1\% mAP and 89.1\% mAP on Market1501 and DukeMTMC-reID, respectively.
To our best knowledge, our baseline achieves best performance in the case of only using global features.

\section{Supplementary Experiments}
We observed that some previous works were done with different the numbers of batch size or image sizes.
In this section, as a supplementary we explore the affects of them on model performance.

\subsection{Influences of the Number of Batch Size}

\renewcommand{\multirowsetup}{\centering}
\begin{table}[tb]\small
  \begin{center}
  \begin{tabular}{ c|cc|cc}
\hline
  Batch Size	 				& \multicolumn{2}{c|}{Market1501} & \multicolumn{2}{c}{DukeMTMC}	 \\
  $P \times K$			& r = 1 	& mAP	&r = 1 	& mAP 	 \\
 	\hline
	\hline
$8\times3$     &92.6	&79.2	&84.4	&68.1			\\
$8\times4$	   &92.9	&80.0	&84.7	&69.4			\\
$8\times6$     &93.5	&81.6	&85.1	&70.7    		\\
$8\times8$     &93.9	&82.0	&85.8	&71.5			\\
$16\times3$    &93.8	&83.1	&86.8	&72.1			\\
$16\times4$    &93.8	&83.7	&86.6	&73.0			\\
$16\times6$    &94.0	&82.8	&85.1	&69.9			\\
$16\times8$    &93.1	&81.6	&86.7	&72.1			\\
$32\times3$    &94.5	&84.1	&86.0	&71.4			\\
$32\times4$	   &93.2	&82.8	&86.5	&73.1			\\
\hline
  \end{tabular}
  \end{center}
  \caption{\label{batchsize}Performance of ReID models with different numbers of batch size.}
\end{table}

The mini-batch of triplet loss includes $B = P \times K$ images.
$P$ and $K$ denote the number of different persons and the number of different images per person, respectively.
A mini-batch can only contain up to 128 images in one GPU, so that we can not do the experiments with $P=32, K=6$ or $P=32, K=8$.
We removed center loss to clearly find the relation between triplet loss and batch size.
The results are present in Table. \ref{batchsize}.
However, there are not specific conclusions to show the effect of $B$ on performance.
A slight trend we observed is that larger batch size is beneficial for the model performance.
We infer that large $K$ helps to mine hard positive pairs while large $P$ helps to mining hard negative pairs.

\subsection{Influences of Image Size}
\renewcommand{\multirowsetup}{\centering}
\begin{table}[htb]\small
  \begin{center}
  \begin{tabular}{ c|cc|cc}
\hline
    	 				& \multicolumn{2}{c|}{Market1501} & \multicolumn{2}{c}{DukeMTMC}	 \\
  Image Size			& r = 1 	& mAP	&r = 1 	& mAP 	 \\
 	\hline
	\hline
$256\times128$ &93.8	&83.7	&86.6	&73.0           \\
$224\times224$ &94.2	&83.3	&86.1	&72.2           \\
$384\times128$ &94.0	&82.7	&86.4	&73.2			\\
$384\times192$ &93.8	&83.1	&87.1	&72.9			\\
\hline
  \end{tabular}
  \end{center}
  \caption{\label{imgsize}Performance of ReID models with different image sizes.}
\end{table}

We trained models without center loss and set $P=16, K=4$.
As shown in Table. \ref{imgsize}, four models achieve similar performances on both datasets.
In our opinion, the image size is not a pretty importance factor for the performance of ReID models.

\section{Conclusions and Outlooks}
In this paper, we collect some effective training tricks and design a strong baseline for person ReID.
To demonstrate the influences of each trick on the performance of ReID models, we do a lot of experiments on both same-domain and cross-domain ReID tasks.
Finally, only using global features, our strong baseline achieve 94.5\% rank-1 accuracy and 85.9\% mAP on Market1501.
We hope that this work can promote the ReID research in academia and industry.

However, the purpose of our work is not to improve performance roughly.
Compared with face recognition, person ReID still has a long way to explore.
We think some training tricks can speed up the exploration and there are many effective tricks not discovered.
We welcome researchers to share some other effective tricks with us.
We will evaluate them based on this work.

In the future, we will continue to design more experiments to analyze the principles of these trciks.
For example, when we replace the BNNeck with L2 normalization, what does the performance of this network become?
In addition, whether can some state-of-the-arts methods such as PCB, MGN and AlignedReID, etc. be expanded on our strong baseline?
More visualization also is helpful for others to understand this work.

\section{Acknowledge}
This work is supported by the National Natural Science Foundation of China (No. 61633019) and the Science Foundation of Chinese Aerospace Industry (JCKY2018204B053).

{\small
\bibliographystyle{ieee_fullname}
\bibliography{egbib}
}

\end{document}